\def\wordcount {1}		

\ifdefined\preprint
  \documentclass[preprint,review,12pt]{elsarticle}
\fi
\ifdefined\wordcount
  \documentclass[final,3p,times,twocolumn]{elsarticle}
\fi
\ifdefined\final
  \documentclass[final,3p,times,twocolumn]{elsarticle}
\fi

\usepackage{graphicx,stfloats}
\usepackage{color}

\usepackage{booktabs}
\usepackage{subcaption}
\usepackage{float}
\usepackage{siunitx}
\usepackage{amsmath,amsfonts,amssymb}
\usepackage{mathrsfs}

\graphicspath{{./figures/}}
\biboptions{sort&compress}

\journal{Proceedings of the Combustion Institute}

\begin{document}

\begin{frontmatter}


\title{Using Physics-Informed Super-Resolution Generative Adversarial Networks for Subgrid Modeling in Turbulent Reactive Flows}

\author[fir]{Mathis Bode\corref{cor1}}
\ead{m.bode@itv.rwth-aachen.de}

\author[sec]{Michael Gauding}
\author[fir]{Zeyu Lian}
\author[fir]{Dominik Denker}
\author[fir]{Marco Davidovic}
\author[fir]{Konstantin Kleinheinz}
\author[thi]{Jenia Jitsev\corref{cor2}}
\author[fir]{Heinz Pitsch\corref{cor2}}

\address[fir]{Institute for Combustion Technology (ITV), RWTH Aachen University, Templergraben 64, 52056 Aachen, Germany}
\address[sec]{CORIA, CNRS UMR 6614, 76801 Saint Etienne du Rouvray, France}
\address[thi]{J\"ulich Supercomputing Centre (JSC), Institute for Advanced Simulation (IAS), Forschungszentrum J\"ulich (FZJ), Wilhelm-Johnen-Stra{\ss}e, 52425 J\"ulich, Germany}

\cortext[cor2]{Equal contribution}
\cortext[cor1]{Corresponding author:}

\begin{abstract}
Turbulence is still one of the main challenges for accurately predicting reactive flows. Therefore, the development of new turbulence closures which can be applied to combustion problems is essential. Data-driven modeling has become very popular in many fields over the last years as large, often extensively labeled, datasets became available and training of large neural networks became possible on GPUs speeding up the learning process tremendously. However, the successful application of deep neural networks in fluid dynamics, for example for subgrid modeling in the context of large-eddy simulations (LESs), is still challenging. Reasons for this are the large amount of degrees of freedom in realistic flows, the high requirements with respect to accuracy and error robustness, as well as open questions, such as the generalization capability of trained neural networks in such high-dimensional, physics-constrained scenarios. This work presents a novel subgrid modeling approach based on a generative adversarial network (GAN), which is trained with unsupervised deep learning (DL) using adversarial and physics-informed losses. A two-step training method is used to improve the generalization capability, especially extrapolation, of the network. The novel approach gives good results in a priori as well as a posteriori tests with decaying turbulence including turbulent mixing. The applicability of the network in complex combustion scenarios is furthermore discussed by employing it to a reactive LES of the Spray A case defined by the Engine Combustion Network (ECN).

\end{abstract}


\end{frontmatter}

\ifdefined \wordcount
\clearpage
\fi

\newcommand{\chist}{\chi_{\mathrm{st}}}
\newcommand{\Zst}{Z_{\mathrm{st}}}
\newcommand{\filt}[1]{\widetilde{#1}}
\newcommand{\Zfilt}{\filt{Z}}
\newcommand{\Zfvar}{\filt{Z''^2}}
\newcommand{\Cfilt}{\filt{C}}
\newcommand{\Cfvar}{\filt{C''^2}}
\newcommand{\rhofilt}{\overline{\rho}}
\newcommand{\Rey}{\mathrm{Re}}

\newcommand{\tsaf}[1]{\langle \filt{#1} \rangle}
\newcommand{\rmsf}[1]{\sqrt{\langle(\filt{#1}-\tsaf{#1})^2 \rangle}}
\newcommand{\means}[1]{{#1}_{\mathrm{mean}}}
\newcommand{\rmss}[1]{{#1}_{\mathrm{RMSD}}}

\newcommand{\di}{\,\mathrm{d}}
\newcommand{\lb}{\left(}
\newcommand{\rb}{\right)}
\newcommand{\ls}{\left[}
\newcommand{\rs}{\right]}

\newcommand{\nl}{ \notag \\ && &}
\newcommand{\nlp}{ \notag \\ && & \quad}
\newcommand{\nle}{ \\ \notag \\ &&}
\newcommand{\nlc}{ + \\ & \quad +}

\newcommand{\reff}[1]{Fig.~\ref{#1}}
\newcommand{\reffl}[1]{Figure~\ref{#1}}
\newcommand{\refe}[1]{Eq.~\eqref{#1}}
\newcommand{\refel}[1]{Equation~\eqref{#1}}
\newcommand{\refs}[1]{Sec.~\ref{#1}}
\newcommand{\refsl}[1]{Section~\ref{#1}}
\newcommand{\refss}[1]{Subsec.~\ref{#1}}
\newcommand{\refssl}[1]{Subsection~\ref{#1}}
\newcommand{\reft}[1]{Table~\ref{#1}}
\newcommand{\reftl}[1]{Table~\ref{#1}}
\newcommand{\refa}[1]{App.~\ref{#1}}
\newcommand{\refal}[1]{Appendix~\ref{#1}}

\newcommand{\rms}{RMSD }

\def\dd#1#2{\frac{\mbox{d} #1}{\mbox{d} #2}}
\def\pp#1#2{\frac{\partial #1}{\partial #2}}
\newcommand{\vect}[1]{\mathbf{#1}}
\newcommand{\vecg}[1]{\mathrm{\textbf{#1}}}
\newcommand{\avg}[1]{\langle{#1}\rangle}
\newcommand{\flu}[1]{{#1}'}
\newcommand{\mat}[1]{\mathbf{#1}}
\newcommand{\tens}[1]{\mathbf{#1}}

\newcommand{\abbrlabel}[1]{\makebox[3cm][l]{\textbf{#1}\ \dotfill}}
\newenvironment{abbreviations}{\begin{list}{}{\renewcommand{\makelabel}{\abbrlabel}}}{\end{list}}

\newcommand{\rebuttal}[1]{{#1}}
\newcommand{\press}[1]{{#1}}
\newcommand{\apress}[1]{{\color{blue}{#1}}}
\newcommand{\picsize}{\small}
\newcommand{\picbox}[1]{{#1}}

\section{Introduction}
\label{dad:sec:introduction}
Machine learning (ML) and deep learning (DL) have gained widespread use and impact in many research communities and industries. The availability of exceptionally large, often extensively labeled datasets and the possibility to train large networks on GPUs, reducing the training time tremendously, are two reasons for this success. Prominent applications of DL include image processing~\cite{dong2014learning,wang2019,greenspan2016guest}, speech recognition~\cite{hinton2012deep}, or learning of optimal complex control ~\cite{Vinyals2019}. These data-driven approaches have been also applied to fluid dynamic problems
~\cite{parish2016paradigm,srinivasan2019predictions,maulik2017neural,bode2018,kutz2017} including works on subgrid modeling for
large-eddy simulation (LES)~\cite{lapeyre2019training,bode2019gan} based on direct numerical simulation (DNS) data. Recently, the idea of physics informed networks~\cite{raissi2019} rose, where architecture or loss function are designed in order to support known properties of the underlying physical problem.
%

Neural networks have been also successfully applied to reactive flows. Examples are the adaptive reduction scheme for modeling reactive flows by Banerjee et al.~\cite{banerjee2006adaptive}, artificial neural network (ANN)-based storage of flamelet solutions~\cite{ihme2009,bode2019bspline}, or direct mapping of LES resolved scales to filtered-flame generated manifolds using customized convolutional neural networks (CNNs) as shown by Seltz et al.~\cite{seltz2019}. Additionally, regularized deconvolution methods, such as published by Wang and Ihme~\cite{wang2019rdm}, are closely related ideas.
%

Often the applications with respect to 
flow data
are limited by either using only simple networks or small, artificial datasets. Thus, many open questions still remain, such as determining proper network architectures for flow problems, searching for hyperparameters, or improving the generalization ability of networks.

This work introduces the application of generative adversarial networks (GANs)~\cite{goodfellow2014generative} for subgrid modeling in turbulent flows\press{, as it seems to be a flexible tool, also promising for reactive turbulent flow simulations.} GANs belong to a particular class of generative models that aim to estimate the unknown probability density that underlies observed data. The special characteristic of this model class is the ability to perform such estimation without an explicitly provided data likelihood function. The learning takes place via an implicit generative model. Learning only requires access to data samples from the unknown distribution. GANs thus perform unsupervised learning of unknown data probability distribution and do not require any labels that are necessary in supervised learning scenarios. Simplified, the particularly interesting feature of GANs is that beside the resulting generator network that is used for modeling, a second network, the discriminator, is used. While the generator creates new modeled data, the discriminator tries to assess, if it is real or generated data and provides that feedback for the training of the generator. Thereby, in each step, the discriminator learns better to identify model versus real data, which will help the generator to generate more realistic model data.
More precisely, estimating an unknown data probability distribution by GAN learning can be understood as {minimax} zero-sum game carried out by two players, the generator and the discriminator, that are both deep networks constituting a full GAN. The game can be described as a generator creating samples to present them to the discriminator, while the discriminator, being confronted with a mix of generated and real data samples, has the task to guess whether a presented input is generated or real data. So, the generator attempts to 'fool' the discriminator, while the discriminator strives to become better in recognizing generated samples from real samples. It was shown that finding the equilibrium of this game corresponds to minimizing different distance measures between the generator model and the true data distribution, such as Kullback-Leibler (KL), Jensen-Shannon (JS) divergence, or Wasserstein distance, depending on a particular form of loss termed adversarial loss. 

Here, a physics-informed enhanced super-resolution GAN (PIESRGAN) is employed, built upon enhanced super-resolution GAN (ESRGAN)~\cite{wang2019} architecture, which has been recently developed in the context of super-resolving GANs (SRGANs)~\cite{ledig2017photo}, to reconstruct fully-resolved turbulence fields from filtered data, such as from LES. To this end, the ESRGAN is extended for three-dimensional (3-D) data handling and, most importantly, endowed with physics-informed loss. Once the fully-resolved data is reconstructed, a filter kernel is applied to close the filtered equations of the LES. \refsl{dad:sec:modeling} describes the PIESRGAN in detail and lies out key features of the network, which are required for an accurate reconstruction. It contains both, a priori and a posteriori tests with decaying turbulence data, including turbulent mixing of a passive scalare, which could be used as mixture fraction in any combustion model. Furthermore, an approach to improve the training and generalization capability, especially for extrapolation, of the trained neural network by combining fully-resolved and under-resolved data is discussed. In \refs{dad:sec:application}, the potential of the novel method is demonstrated by using PIESRGAN as subfilter model for the filtered momentum and scalar equations in an LES of the Spray A case defined by the Engine Combustion Network (ECN)~\cite{ecn2019}, which is a complex reactive turbulent flow featuring high Reynolds numbers and spray. The paper finishes with conclusions and future work.

\section{Modeling}
\label{dad:sec:modeling}
%
A subgrid model needs to predict the subfilter statistics of fully-resolved data (e.\,g. DNS data denoted with 'H'), knowing only the corresponding filtered data with reduced information content (e.\,g. LES data denoted with 'F'). Here, the fully-resolved data $\phi_\mathrm{H}$ and the filtered data $\phi_\mathrm{F}$ are connected by a filter operation $\phi_\mathrm{F} = \mathscr{F} (\phi_\mathrm{H})$, for example with a Gaussian filter kernel. The filtered equations, which are solved in LES, could be also closed if the fully-resolved data is reconstructed with an inverse filter operation $\phi_\mathrm{H} = \mathscr{F}^{-1} (\phi_\mathrm{F})$ that statistically restores the original, fully-resolved data.

The described challenges of subgrid models are similar to challenges faced in super-resolution imaging. Here, SRGANs were found to be a successful tool for approximating the inverse deconvolution operator $\phi_\mathrm{R} = \tilde{\mathscr{F}}^{-1} (\phi_\mathrm{F}) \approx \mathscr{F}^{-1} (\phi_\mathrm{F})$~\cite{wang2019,ledig2017photo}, where $\phi_\mathrm{R}$ denotes the reconstructed, high-resolution data.
Thus, PIESRGAN is used as approximation $\tilde{\mathscr{F}}^{-1}$. For example, if $\phi_\mathrm{F}^k$ denotes a filtered solution at time $k$, the resulting simulation workflow closing a chemical source term reads:
\begin{enumerate}
\item Use the PIESRGAN to reconstruct $\phi_\mathrm{R}^k$ from $\phi_\mathrm{F}^k$.
\item Use $\phi_\mathrm{R}^k$ to estimate the unclosed terms $\dot{\omega}_\mathrm{F}$ in the filtered transport equation of $\phi$ by evaluating the local source term with $\phi_\mathrm{R}^k$ and applying a filter operator.
\item Use $\dot{\omega}_\mathrm{F}$ and $\phi_\mathrm{F}^k$ to advance the filtered transport equation of $\phi$ to $\phi_\mathrm{F}^{k+1}$.
\item Repeat 1.--3.
\end{enumerate}

\subsection{Network architecture}
\label{dad:ssec:network}
\begin{figure*}[h]
    \centering
    \includegraphics[width=144mm]{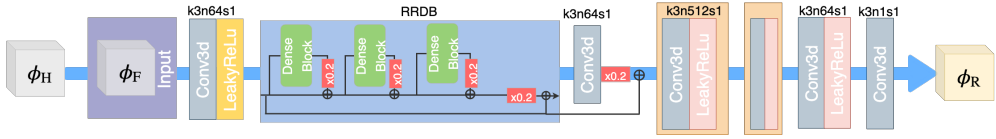}
    \caption{Generator architecture of PIESRGAN.}
    \label{dad:fig:generator}
\end{figure*}

The generator of the PIESRGAN is depicted in \reff{dad:fig:generator}. It is fed with 3-D subboxes of the flow fields during training and heavily uses
3-D CNN layers (Conv3D)~\cite{krizhevsky2012imagenet} in combination with leaky rectified linear unit~(LeakyReLU) layers for activation. The convolutional layers are capable of extraction increasingly complex multi-dimensional features with increasing network depth.
%
%
%

The residual in residual dense block~(RRDB), which is introduced in ESRGAN and replaces the residual block~(RB) used in previous architectures, is important for the performance of state-of-the-art SRGANs. The RRDB contains fundamental architectural elements, such as residual dense blocks (DBs) with skip-connections, where in turn each DB uses dense connections inside. The output from each layer within the DB is sent to all the following layers. For PIESRGAN, DBs are repeated three times, using residual skip connections as shown in \reff{dad:fig:generator} with the residual scaling factor $\beta_{\mathrm{RSF}}=0.2$, which helps to avoid instabilities in the forward and backward propagation. The motivation behind the RRDB architecture is to enable generation of super-resolved data through a very deep network that has enough capacity to learn and model all relevant complex transformations that are necessary to specify the required reconstruction operation.
%

As suggested by Wang et al.~\cite{wang2019}, all batch normalization~(BN) layers of the ESRGAN architecture were removed for PIESRGAN. This reduces the computational cost bound to BN and was shown to improve the performance with respect to former single image super-resolution (SISR) models that utilized BN~\cite{lim2017enhanced}. Moreover, using BN layers was shown to introduce distorting artifacts into generated images, which is absolutely undesirable for modeling turbulence.
%

Another difference of PIESRGAN to traditional SISR applications lies in the input and output dimensions. In SISR, where the generated high resolution image contains an increased number of pixels, the fully-resolved data in turbulence contain finer structures that are enclosed in the flow. Therefore, turbulence super-resolution does not involve classical upsampling or downsampling. The input and output hence hold the same dimension, while the output flow has more energy distributed in the high wave number part.
%
%

\begin{figure*}[h]
    \centering
    \includegraphics[width=144mm]{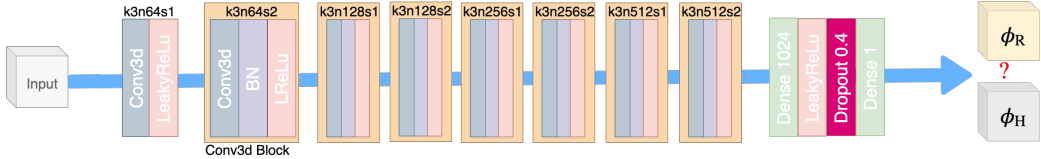}
    \caption{Discriminator architecture of PIESRGAN.}
    \label{dad:fig:dis}
\end{figure*}

The discriminator inherits the basis CNN architecture as shown in \reff{dad:fig:dis}. It consists of one Conv3d block without BN and seven Conv3d blocks with BN, followed by a fully connected layer block with DropOut. LeakyReLu layers are used for activation.
The blocks close to the input learn relative simple features extracted from turbulent flows, whereas the blocks close to the output learn more complex, high-level features, like eddies/vortexes. The number of filter maps increases with depth following conventional design. The DB starts with a Dense(1024) hidden layer, which projects highly dimensional output from many filter maps of the final Conv3d block into a flat 1024 dimensional vector. The following Droupout layer with factor $0.4$ serves as regularization, reducing the risk of overfitting. A relativistic adversarial loss as proposed by Jolicoeur-Martineau~\cite{jolicoeur2018relativistic} is used. Using relativistic loss as adversarial loss was shown to stabilize GAN training in different scenarios~\cite{jolicoeur2018relativistic}. It also presumably aids learning of sharper edges and more detailed textures in SISR cases, which should also help to learn very high frequency details in the turbulence context. 

The perceptual loss proposed for the ESRGAN based on the VGG-feature space pretrained with ImageNet dataset is less suitable for turbulence data, as the natural image features from VGG19 may be not representative for turbulent flows. Instead, physics-informed constraints are incorporated into the loss function, guided by laws governing the physics of turbulence flow. More precisely, the loss function for PIESRGAN is chosen as 
\begin{equation}
\mathcal{L} = \beta_1 L_\mathrm{adv} + \beta_2 L_\mathrm{pixel} + \beta_3 L_\mathrm{gradient} + \beta_4 L_\mathrm{continuity},
\end{equation}
where $\beta_1$, $\beta_2$, $\beta_3$, and $\beta_4$ are coefficients weighting the different loss term contributions. $L_\mathrm{adv}$ is the discriminator/generator relativistic adversarial loss~\cite{wang2019}, which reflects both how well the generator is able to generate high resolution turbulence samples that look like real, DNS-obtained full-resolved turbulence flows and how well the discriminator is still able to recognize real and generated flows apart. The pixel loss $L_\mathrm{pixel}$ and the gradient loss $L_\mathrm{gradient}$ are defined as mean-squared error (MSE) of the quantity itself and of the gradient of the quantity, respectively~\cite{bode2019gan}. If the MSE operator is applied on tensors including vectors, such as the velocity, it is applied to all elements separately.
Afterwards the resulting tensor is mapped into a scalar using the $L_1$-norm. $L_\mathrm{continuity}$ is the physics-informed continuity loss, which contributes to the total loss enforcing those physically plausible solutions of the reconstructed flow field where divergence of the velocity field should be zero for incompressible flows.
If no reference solution exists, $\beta_2$ and $\beta_3$ are set to zero, reducing the loss to adversarial loss and potential continuity loss.

\subsection{Implementation details}
\label{dad:ssec:implementation}
All networks were trained
using cropped sub-boxes with size $16\times16\times16$ from the DNS and the corresponding filtered low-resolution flow field. This box size was found to be a good compromise between memory requirement during the reconstruction step and the characteristic length scales of the flow and filter width. For mapping a passive scalar field combined with the velocity fields, each batch with batch size 32, which is the number of samples processed before the model is updated, has the dimension $32\times 16\times 16\times 16 \times 4$, where four channels consist of one passive scalar channel and three velocity channels. The flow field at a given time step was divided into non-overlapping sub-boxes, which were all used for training in one epoch and accessed in random order. RMSProp, which relies on the stochastic gradient descent (SGD) approach, was used as optimizer.
%
%
All fields were zero mean-centered and rescaled with the variance before using them for training.

In order to increase the reproducibility of this work and clarify more technical details, the implemented PIESRGAN was uploaded to GIT (https://git.rwth-aachen.de/Mathis.Bode/PIESRGAN.git).

\subsection{Training strategy}
\label{dad:ssec:trainings}
Many industrially relevant applications are operated at very high Reynolds numbers that are not accessible by DNS.
Thus training the network only with DNS data of the relevant Reynolds number range is not possible. This raises the question whether a network trained with DNS data of lower Reynolds numbers is general enough to give also good results at higher Reynolds numbers, i.\,e. whether it has an extrapolation capability.

It will be shown in the a priori test that training the network only with DNS data leads to bad accuracy for Reynolds numbers outside of the training range. Therefore, the training is extended by a second step in this work. After training generator and discriminator simultaneously with DNS data ('H') and corresponding filtered data ('F'), the generator is further trained and updated using filtered data ('$\tilde{\mathrm{F}}$'), which were generated for a larger Reynolds number range with LES without subfilter closing, which can be computed at low computational cost. Corresponding 'H' data do not exist, and the discriminator is not further updated.
Note that the loss function reduces for this second learning phase as already mentioned before, as the evaluation of loss terms related to DNS data is no longer possible. Thus, loss is driven mainly by the part of adversarial loss that corresponds to correctly recognizing generated flow samples and by the physics-informed continuity constrain. 

\subsection{A priori testing}
\label{dad:ssec:apriori}
\begin{figure*}[!htb]
    \centering
        \begin{subfigure}[b]{27mm}
            \centering
            \includegraphics[width=\textwidth]{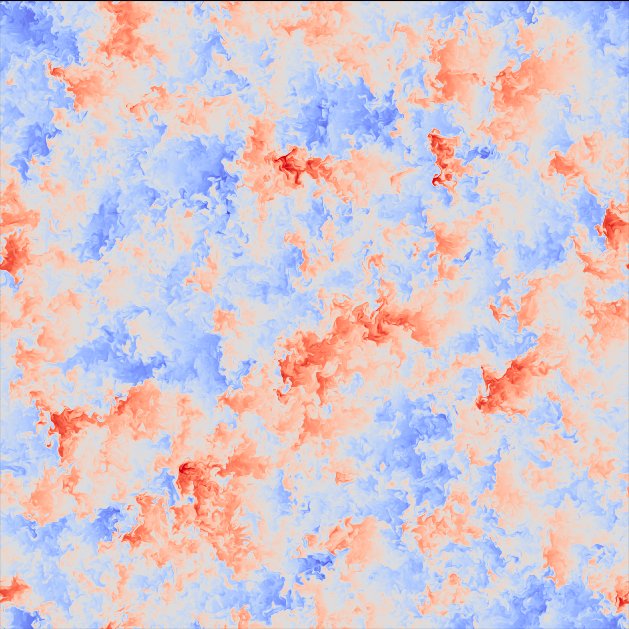}
            \vspace*{-10mm}
            \caption*{\colorbox{white}{$z_{\mathrm{H}}$}}    
            \label{dad:sfig:dns:z}
        \end{subfigure}
        \hspace{1mm}
        \begin{subfigure}[b]{27mm}
            \centering
            \includegraphics[width=\textwidth]{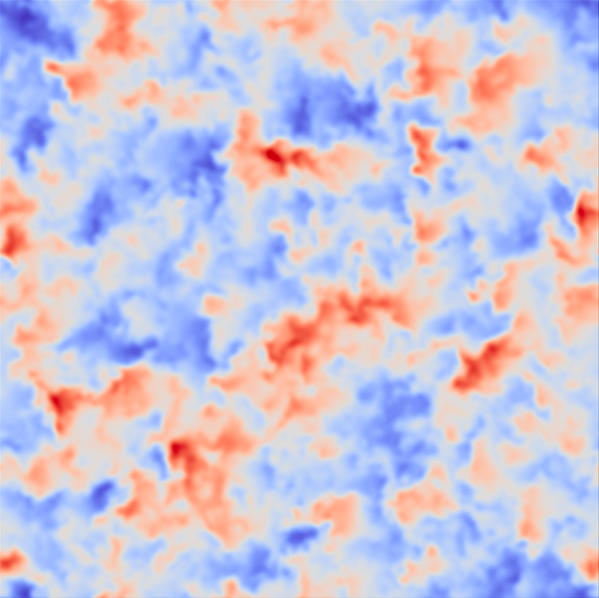}
            \vspace*{-10mm}
            \caption*{\colorbox{white}{$z_{\mathrm{F}}$}}   
            \label{dad:sfig:filt:z}
        \end{subfigure}
        \hspace{1mm}
        \begin{subfigure}[b]{27mm}
            \centering
            \includegraphics[width=\textwidth]{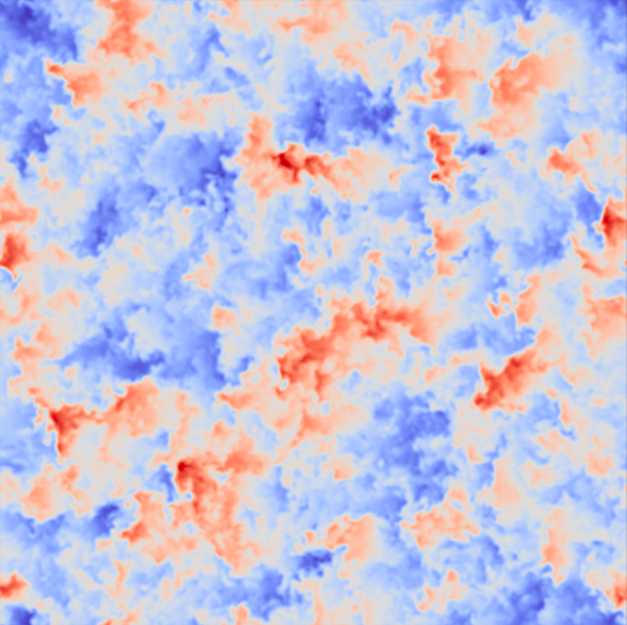}
            \vspace*{-10mm}
            \caption*{\colorbox{white}{$z_{\mathrm{R}'}$}}   
            \label{dad:sfig:recc:z}
        \end{subfigure}
        \hspace{1mm}
        \begin{subfigure}[b]{27mm}
            \centering
            \includegraphics[width=\textwidth]{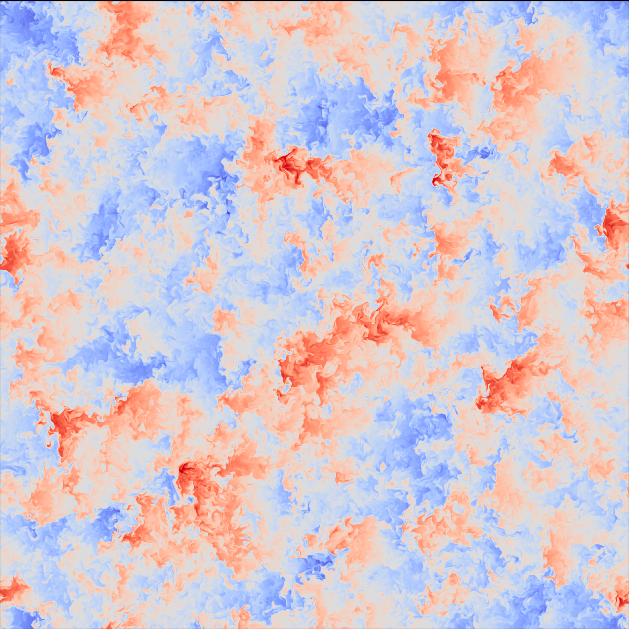}
            \vspace*{-10mm}
            \caption*{\colorbox{white}{$z_{\mathrm{R}}$}}   
            \label{dad:sfig:rec:z}
        \end{subfigure}
        \hspace{1mm}
        \begin{subfigure}[b]{27mm}
            \centering
            \includegraphics[width=\textwidth]{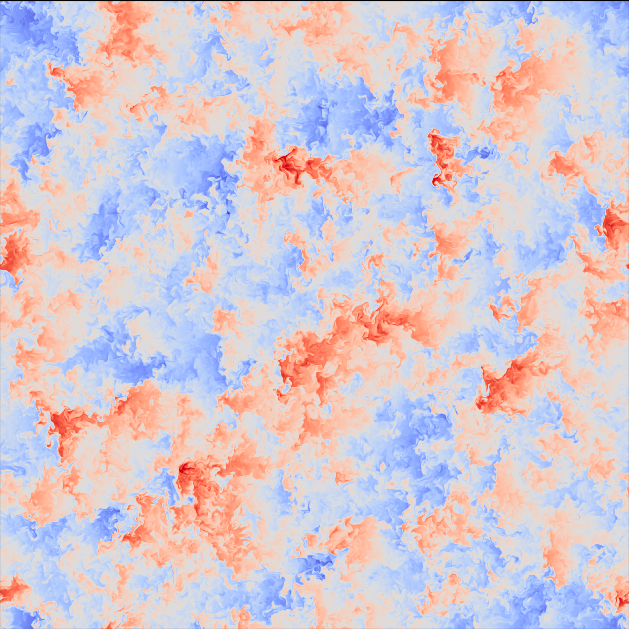}
            \vspace*{-10mm}
            \caption*{\colorbox{white}{$z_{\mathrm{R}''}$}}   
            \label{dad:sfig:recl:z}
        \end{subfigure}    
    \vskip 1mm
    
        \begin{subfigure}[b]{27mm}
            \centering
            \includegraphics[width=\textwidth]{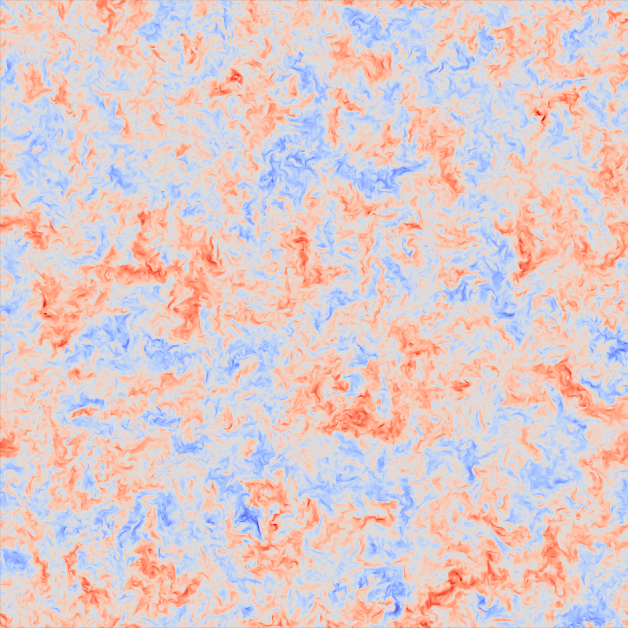}
            \vspace*{-10mm}
            \caption*{\colorbox{white}{$u_\mathrm{H}$}}    
            \label{dad:sfig:dns:u}
        \end{subfigure}
        \hspace{1mm}
         \begin{subfigure}[b]{27mm}
            \centering
            \includegraphics[width=\textwidth]{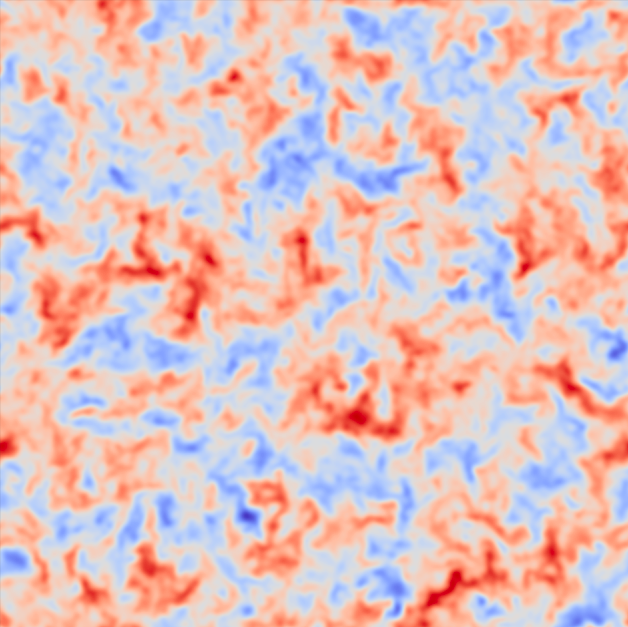}
            \vspace*{-10mm}
            \caption*{\colorbox{white}{$u_\mathrm{F}$}}      
            \label{dad:sfig:filt:u}
        \end{subfigure}      
        \hspace{1mm}
        \begin{subfigure}[b]{27mm}
            \centering
            \includegraphics[width=\textwidth]{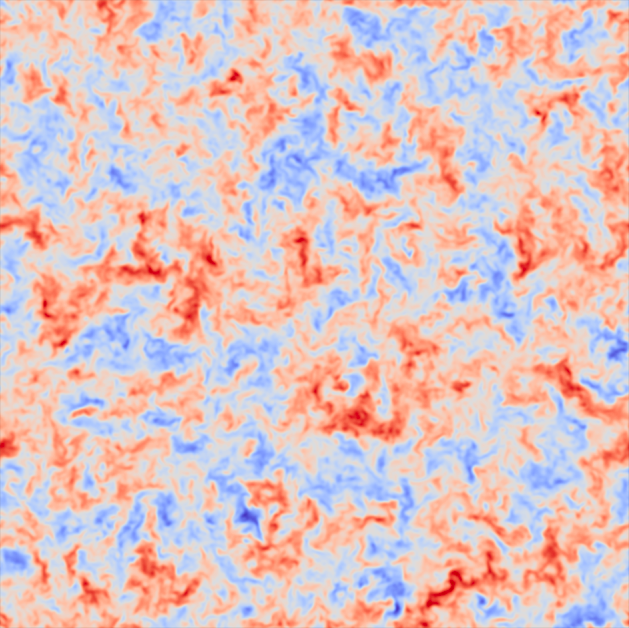}
            \vspace*{-10mm}
            \caption*{\colorbox{white}{$u_{\mathrm{R}'}$}}     
            \label{dad:sfig:recc:u}
        \end{subfigure}
        \hspace{1mm}
        \begin{subfigure}[b]{27mm}
            \centering
            \includegraphics[width=\textwidth]{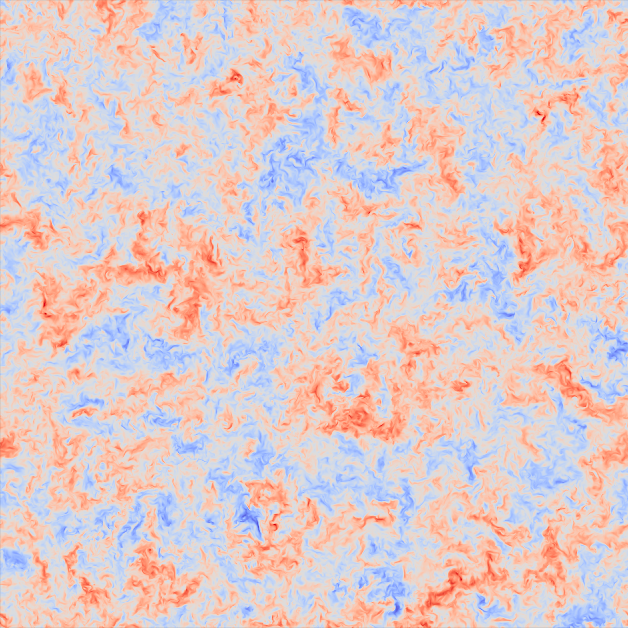}
            \vspace*{-10mm}
            \caption*{\colorbox{white}{$u_{\mathrm{R}}$}}    
            \label{dad:sfig:rec:u}
        \end{subfigure}
        \hspace{1mm}
        \begin{subfigure}[b]{27mm}
            \centering
            \includegraphics[width=\textwidth]{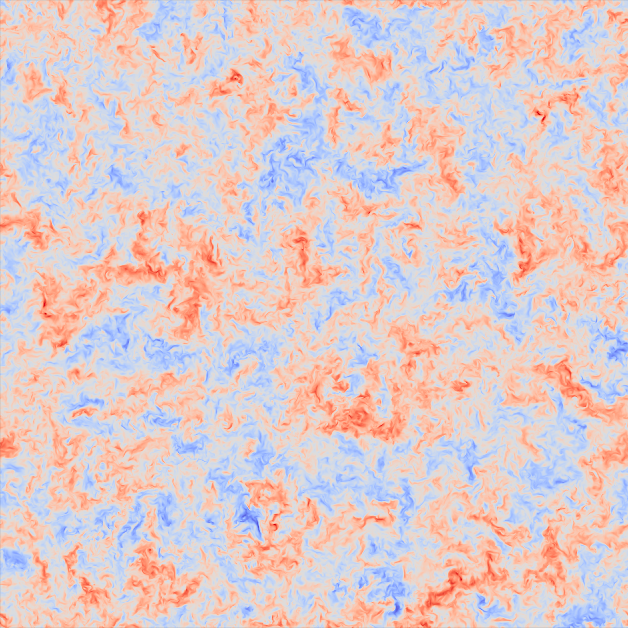}
            \vspace*{-10mm}
            \caption*{\colorbox{white}{$u_{\mathrm{R}''}$}}    
            \label{dad:sfig:recl:u}
        \end{subfigure}
       \caption{Visualization of 2-D slices of the fluctuations of the passive scalar $z$ and of the velocity component $u$ for the time step with Taylor microscale-based Reynolds number of about 88.}
        \label{dad:fig:vis}
    \end{figure*}

One of the largest existing decaying turbulence DNS datasets~\cite{gauding2019self} was used for training and testing the PIESRGAN. The dataset features periodic boxes of homogeneous isotropic turbulence with Reynolds numbers based on the Taylor microscale of up to 88, which were simulated on $4096^3$ mesh points. The first data time step is defined to lie in the self-similar range of the flow, as indicated with $t_\mathrm{start}$ in \reff{dad:fig:evo}. Before the training, the data was filtered to obtain combinations of 'H' and 'F'. The PIESRGAN was able to reconstruct data within the trained Reynolds number range well. For testing the extrapolation capability of the network, the first time step of the DNS data was not used for training but only for testing. As the Reynolds number reduces over time for the decaying turbulence case, skipping the first time step of the data resulted in a highest Reynolds number for training of about 75, while testing was performed with a Reynolds number of 88. The results are shown in \reff{dad:fig:vis} as '$\mathrm{R}'$' for the fluctuation of a passive scalar $z$ as well as the fluctuation of one velocity component, which is labeled as $u$. Obviously, the network adds insufficient small scale structures to the flow. Maybe because it had never seen such a high Reynolds number before, i.e. never needed to add so small structures to the flow. The column labeled with 'R' shows the results of a network additionally trained with '$\tilde{\mathrm{F}}$' data, featuring Reynolds numbers based on the Taylor microscale of up to 250. The reconstruction results are much better and the visual agreement with the DNS data is almost perfect. One reason for this could be that the '$\tilde{\mathrm{F}}$' data just modifies all weights in the network, which results in more subfilter contributions for all Reynolds numbers, randomly leading to the good reproduction for the target Reynolds number but worse results for the others. That would contradict the idea that the neural network used the new data to really learn the target Reynolds number results by means of the adversarial loss. Therefore, the PIESRGAN was alternatively trained with the 'H'/'F' dataset and additionally a dataset '$\tilde{\mathrm{F}'}$', featuring only Taylor microscale-based Reynolds number of up to 200. These results are shown in \reff{dad:fig:vis} as '$\mathrm{R}''$', and the agreement is similarly good as before, which indicates that the network really learned to reproduce the higher Reynolds number data. Without showing the results here, this was also emphasized by analyzing the results for '$\mathrm{R}'$' and '$\mathrm{R}$' in the Reynolds number training range, which did not differ. The same result was also observed for the other two velocity components, which are not visualized in \reff{dad:fig:vis}.

\begin{figure}[!htb]
\picsize
	\centering
    \picbox{\input{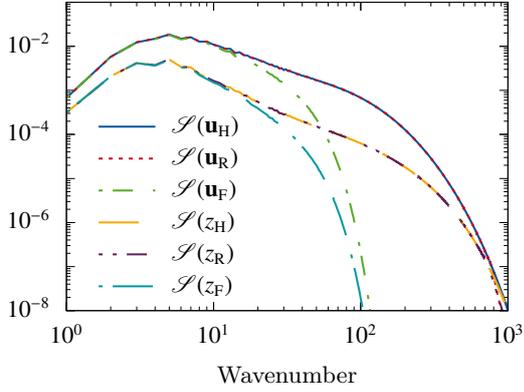}}
	\caption{Spectra evaluated on DNS data, filtered data, and reconstructed data for the velocity vector $\vect{u}$ and the passive scalar $z$ for the time step with Reynolds number of about 88.}
	\label{dad:fig:spe}
\end{figure}

In addition to the visual evaluation, \reff{dad:fig:spe} presents the spectrum denoted with $\mathscr{S}$ computed with 'H', 'F', and 'R' data. It shows that also the statistical agreement between DNS and reconstructed data is very good. Only for very high wavenumbers, the reconstructed flow field slightly differs from the DNS data. Note that the spectrum based on the velocity uses all three velocity components. Therefore, $\mathscr{S}(\vect{u})$ with bold notation for vectors is shown.

The results presented in \reff{dad:fig:vis} and \reff{dad:fig:spe} indicate that the PIESRGAN is able to learn universal key features of turbulence with the adversarial loss, which enable the correct prediction of statistics of higher Reynolds number flows, only seeing filtered data. This is a big advantage to simpler networks fully relying on supervised learning. How the network is able to detect the target Reynolds number from the provided fluctuation is an open question and should be addressed in more detail in future work.

\subsection{A posteriori testing}
\label{dad:ssec:aposteriori}
\begin{figure}[!htb]
\picsize
	\centering
    \picbox{\input{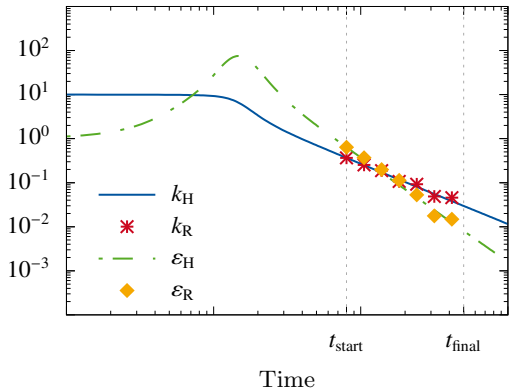}}
	\caption{Temporal evolution of the ensemble-averaged turbulent kinetic energy $k$ and ensemble-averaged dissipation rate $\varepsilon$.}
	\label{dad:fig:evo}
\end{figure}

Before using the trained network in a complex reactive turbulent flow, an a posteriori test is performed with respect to the decaying turbulence data. For that, filtered data of the early time step $t_\mathrm{start}$ of the decaying turbulence DNS case are used as initial flow field and advanced over time according to the steps outlined at the beginning of this section. In order to keep the filter width of the data consistent to the training data, the DNS data of size $4096^3$ are filtered to a $64^3$ mesh. The time step size of the LES was increased compared to the DNS. \reffl{dad:fig:evo} compares the decay of the ensemble-averaged turbulent kinetic energy $k$ and the ensemble-averaged dissipation rate $\varepsilon$ evaluated during the DNS and the a posteriori test with PIESRGAN as LES model. The good agreement between DNS and PIESRGAN-LES is remarkable. During the decay, the Kolmogorov length-scale and the integral length-scale increase with time following a power-law. This implies that the number of wavenumbers that need closure decreases during the decay. The PIESRGAN accounts for this change of the relative relevance of the subgrid closure, which underlines its ability to model small-scale turbulence.

\section{Application}
\label{dad:sec:application}
\begin{figure}[!htb]
\picsize
	\centering
    \picbox{\includegraphics[width=50mm]{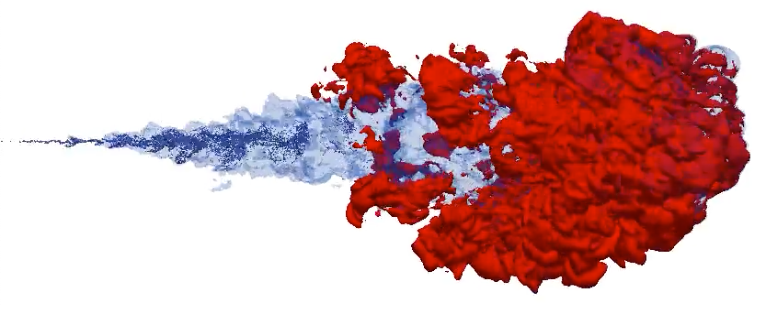}}
	\caption{Visualization of the LES with PIESRGAN as subgrid model for the mixture fraction and velocity applied to the Spray A case: Liquid droplets (dark blue), stoichiometric mixture fraction (light blue), and iso-temperature-surface \SI{1100}{\kelvin} (red). A video of the injection can be found here: https://youtu.be/86gZEhRB5oY.}
	\label{dad:fig:spraya}
\end{figure}

\vspace{-2mm}
\enlargethispage*{\baselineskip}
One prominent example for turbulent reactive flows is the Spray A case (Taylor microscale-based Reynolds

\noindent numbers of up to 235) defined by the ECN~\cite{ecn2019}, which is chosen to demonstrate the usage of the trained PIESRGAN for combustion here. PIESRGAN is used as LES-subgrid model for the subfilter turbulent flux in the mixture fraction equation and for the subfilter Reynolds-stresses in the momentum equations. These quantities are known to be important for accurate results~\cite{davidovic2017}. More precisely, the same conditions and simulation setup as in Davidovic et al.~\cite{davidovic2017} are computed, using the chemical mechanism of Yao et al.~\cite{yao2017} as well as a multiple representative interactive flamelets (MRIF) model. Details of the simulation setup and numerics can be found in former publications~\cite{davidovic2017,bode2015a,bode2017a,bode2019tcp}.  Compared to the simulations performed by Davidovic et al.~\cite{davidovic2017}, a coarser mesh was used in this work to emphasize effects of the subgrid model, resulting in a minimum grid spacing of \SI{100}{\micro\meter} close to the nozzle. The LES result with PIESRGAN as subgrid model for mixture fraction and velocity is visualized in \reff{dad:fig:spraya}. Note that the PIESRGAN was also used to evaluate the mixture fraction variance on the reconstructed mixture fraction field. It could have also been used for computing the probability density function (PDF), used as part of the MRIF model, but instead a classical beta-PDF was used here. Furthermore, the ignition delay time, defined as the time, when the OH mass fraction reaches \SI{2}{\%} of its maximum value for the first time, was evaluated as \SI{0.379}{\milli\second} (averaged over four realizations), which is in reasonable agreement to about \SI{0.4}{\milli\second} measured in experiments~\cite{ecn2019}.

\begin{figure}[!htb]
\picsize
	\centering
    \picbox{\input{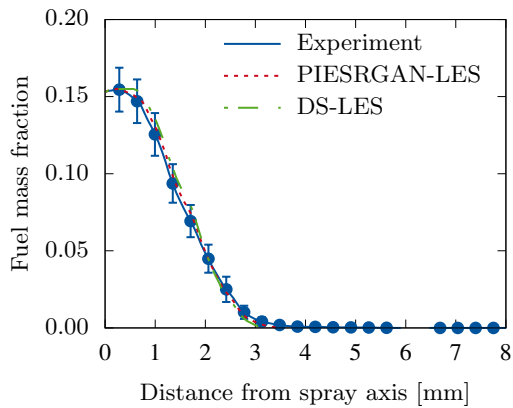}}
	\caption{Temporally and circumferentially averaged fuel mass fraction evaluated \SI{18.75}{\milli\metre} downstream from the nozzle.}
	\label{dad:fig:mixing}
\end{figure}

In order to assess the effect of the subgrid modeling on the mixing, the fuel mass fraction is evaluated \SI{18.75}{\milli\metre} downstream from the nozzle. It is temporally and circumferentially averaged and plotted in \reff{dad:fig:mixing} for the PIESRGAN-LES, an LES with dynamic Smagorinsky model~\cite{davidovic2017} (denoted with 'DS-LES'), and experimental data~\cite{pickett2015}. The mixing of the PIESRGAN-LES lies in between the experimental data and the DS-LES results. This indicates that the PIESRGAN is a robust and accurate model, which is interesting considering that it was trained with homogeneous isotropic turbulence data. Note that the PIESRGAN-LESs were run without any clipping, which weakens the hypothesis that data-driven models are dangerous to use in real simulations as extreme predictions might crash the simulation. Computationally, the PIESRGAN-LES was more expensive than the DS-LES. However, with the rapid improvements in the field of DL on GPUs, this could change in the near future.

\section{Conclusions}
\label{dad:sec:conclusions}
This work presents a novel GAN-based subgrid modeling approach, which employs unsupervised DL with a combination of super-resolution adversarial and physics-informed losses for accurately predicting subfilter statistics in a wide Reynolds number range. \press{The PIESRGAN is trained with some of the largest existing decaying turbulence data. A successive training with fully-resolved and under-resolved data increases the generalization capability of the network.} It is shown that the trained network gives good results in a priori as well as a posteriori tests and even in a reactive LES with spray. Even though some aspects of the network are not fully understood yet and the data processing speed needs to be improved, this work emphasizes the large potential of data-driven models for reactive flows.

The GAN-method was applied for modeling subfilter terms for momentum and scalar mixing in this work. However, the application to reactive scalar fields to close the chemical source term might be interesting. This is challenging for different reasons. For instance, for fast chemistry, the source term depends on the very smallest scales, which means that these need to be correctly predicted for multi-scalar fields. Still, it will be interesting to assess the potential in future work. 

\bibliography{bib/dad} 
\bibliographystyle{elsarticle-num-PROCI.bst}


\end{document}